\documentclass[runningheads]{llncs}
\usepackage{graphicx}
\usepackage{amsmath,amssymb} 
\usepackage{color}
\usepackage[width=122mm,left=12mm,paperwidth=146mm,height=193mm,top=12mm,paperheight=217mm]{geometry}

\usepackage{epsfig}
\usepackage{psfrag}

\usepackage{xcolor}

\DeclareMathOperator*{\argmin}{arg\,min}
\DeclareMathOperator*{\shrink}{shrink}
\newcommand{\repeatthanks}{\textsuperscript{\thefootnote}}

\newcommand{\R}{\mathbb{R}}

\newcommand{\erdosrenyi}{Erd\H{o}s-R\'{e}nyi\ }

\newcommand{\tnot}{t^{(0)}}

\newcommand{\tnotbar}{\bar{t}^{(0)}}
\newcommand{\toi}{t^{(0)}_{i}}

\newcommand{\toj}{t^{(0)}_{j}}

\newcommand{\vij}{v_{ij}}

\newcommand{\Eb}{E_b}
\newcommand{\proj}{P}
\newcommand{\Pvijp}{\proj_{\vij^\perp}}

\newcommand{\tonum}[1]{t^{(0)}_{#1}}

\newcommand{\Tnot}{T^{(0)}}

\begin{document}
\pagestyle{headings}
\mainmatter

\title{ShapeFit and ShapeKick \\ for Robust, Scalable Structure from Motion} 

\titlerunning{ShapeFit and ShapeKick for Robust, Scalable Structure from Motion}

\authorrunning{Goldstein, Hand, Lee, Voroninski, and Soatto}

\author{Thomas Goldstein$^a$\thanks{These authors contributed equally}, Paul Hand$^b$\repeatthanks, Choongbum Lee\repeatthanks, Vladislav Voroninski$^{c}$\repeatthanks, and Stefano Soatto$^d$}


\institute{$^a$Dept. of Computer Science, University of Maryland, College Park, MD,\\
$^b$Dept. of Computational and Applied Mathematics, Rice University, Houston, TX,\\
$^c$Dept. of Mathematics, Massachusetts Institute of Technology, Cambridge, MA,\\
$^d$Dept. of Computer Science, University of California, Los Angeles, CA.
	\email{tomg@cs.umd.edu, hand@rice.edu, vlad@helm.ai, soatto@ucla.edu}
}

\maketitle

\begin{abstract}
   We introduce a new method for location recovery from pairwise directions that leverages an efficient convex program that comes with exact recovery guarantees, even in the presence of adversarial outliers. When pairwise directions represent scaled relative positions between pairs of views (estimated for instance with epipolar geometry) our method can be used for location recovery, that is the determination of relative pose up to a single unknown scale. For this task, our method yields performance comparable to the state-of-the-art with an order of magnitude speed-up. Our proposed numerical framework is flexible in that it accommodates other approaches to location recovery and can be used to speed up other methods. These properties are demonstrated by extensively testing against state-of-the-art methods for location recovery on 13 large, irregular collections of images of real scenes in addition to simulated data with ground truth. 

\end{abstract}

\section{Introduction}
The typical structure-from-motion (SfM) pipeline consists of (i) establishing sparse correspondence between local regions in different images of a (mostly) rigid scene, (ii) exploiting constraints induced by epipolar geometry to obtain initial estimates of the relative pose (position and orientation) between pairs or triplets of views from which the images were captured, where each relative position is determined up to an arbitrary scale, (iii) reconciling all estimates and their scales to arrive at a consistent estimate up to a single global scale, finally (iv) performing bundle adjustment to refine the estimates of pose as well as the position of the sparse points in three-dimensional (3D) space that gave rise to the local regions in (i), also known as feature points.

As in any cascade method,\footnote{The standard pipeline stands in opposition to direct methods that minimize the discrepancy between the measured images and the images predicted by a forward rendering model with respect to the (infinite-dimensional) shape of the scene, which gives rise to a variational optimization problem which we do not address here.} the overall solution is sensitive to failures in the early stages. While significant effort has gone into designing better descriptors for use in stage (i) of the pipeline, sparse correspondence is intrinsically local and therefore subject to ambiguity. This forces subsequent stages (ii), (iii) to deal with inevitable correspondence failures, often by solving combinatorial matching problems. Stages (ii) and (iv) are well established and are the subject of textbooks. Thus, we hone in on the weak link of the pipeline (iii) to {\em develop global alignment methods that are robust to failure of the correspondence stage.} Towards this end, we propose a novel efficient approach based on convex optimization that comes with provable recovery guarantees. 

Errors in the correspondence stage (i) usually come in two distinct flavors.  First, localization error due to quantization artifacts and sensor noise, which can be modeled as independently and identically-distributed (i.i.d.) additive perturbations drawn from a normal density with zero mean and constant covariance. Second, mismatches due to gross violations of the assumptions underlying local correspondence: co-visibility, constant illumination, and rigidity. The latter can also be modeled as an additive (non i.i.d.) perturbation with unknown distribution. Sparse correspondence errors that arise from only the first source of error, often referred to as ``noise,'' are called {\em inliers}, whereas those subject to both are {\em outliers}. Following a classical robust statistical approach, we forgo modeling the distribution of outliers, and indeed allow them to behave in an {\em adversarial} manner.  We seek algorithms with provable guarantees despite such behavior, while simultaneously being efficiently solvable with low complexity numerical methods.

\subsection{Related work and contributions}

There is a vast literature on sparse matching (i), epipolar geometry (ii) and bundle adjustment (iv) for which we refer the reader to standard Computer Vision textbooks.  Stage (iii) can be separated into two parts: global rotation estimation and location recovery.  For simplicity, we assume that the intrinsic calibration parameters of all cameras are known.

There are many efficient and stable algorithms for estimating global camera rotations \cite{1dSfm_3,1dSfM_5,1dSfm_7,1dSfm_8,1dSfm_10,1dSfm_11,AMIT_13,1dSfm_13,1dSfm_17,1dSfm_18,AMIT_25,1dSfm_19}.  Empirically, \cite{tron2016survey} demonstrates that a combination of  filtering, factorization, and local refinement can accurately estimate 3d rotations.
Theoretically, \cite{WangSinger} prove that rotations can be exactly and stably recovered for a synthetic model by a least unsquared deviation approach on a semidefinite relaxation.  Alternatively, in many applications, such as location services from mobile platforms, augmented reality, and robotics, orientation can be estimated far more reliably than location and scale due to the relatively small gyrometer bias compared to the doubly-integrated accelerometer bias and global orientation references provided by gravity and magnetic field.

We concentrate on the {\em location recovery} problem from relative directions based on known camera rotations.
There have been many different approaches to this problem, such as least squares \cite{1dSfm_3,1dSfm_4,1dSfm_11,1dSfm_17}, second-order cone programs, $l_\infty$ methods \cite{1dSfm_16,AMIT_19,1dSfm_17,1dSfm_18,AMIT_27}, spectral methods \cite{1dSfm_4}, similarity transformations for pair alignment \cite{1dSfm_19}, Lie-algebraic averaging \cite{AMIT_13}, Markov random fields \cite{1dSfm_6}, and several others \cite{1dSfm_14,AMIT_25,1dSfm_19,AMIT_32}. Unfortunately, many location recovery algorithms either lack robustness to mismatches, at times produce collapsed solutions \cite{AMIT_25}, or suffer from convergence to local minima, in sum causing large errors in (or even complete degradation of) the recovered locations.

Recent advances have addressed some of these limitations: 1dSfM \cite{1dsfm} focuses on removing outliers by examining inconsistencies along one-dimensional projections, before attempting to recover camera locations. This method, however, does not reason about self-consistent outliers, which can occur due to repetitive structures, commonly found in man-made scenes.  Also, Jiang et al. \cite{jiang2013global} introduced a method to filter outlier epipolar geometries based on inconsistent triplets of views.   \"{O}zye\c{s}\.{i}l and Singer propose a convex program called Least Unsquared Deviations (LUD) and empirically demonstrate its robustness to outliers \cite{AMIT}.  While these methods exhibit {good} empirical performance, they lack theoretical guarantees in terms of robustness to outliers.

\textbf{Summary of contributions.} In this paper, we propose a novel framework for location recovery from pairwise direction observations.  This framework, called ShapeFit, is based on convex optimization and can be proven to recover locations \textit{exactly} in the presence of adversarial corruptions, under rather broad technical assumptions.  We introduce two efficient numerical implementations, ShapeFit and ShapeKick (both described in Sect. \ref{sect:numerics}), show how they can be employed to solve location recovery problems arising in SfM problem with known camera rotations, and extensively validate our methods using benchmark datasets (Sect. \ref{sect-expm}) and show that our approach achieves significant computational speedups at comparable accuracy. 

\subsection{Problem formulation}

Let $T$ be a collection of $n$ distinct vectors $\tonum{1},\tonum{2},\ldots,\tonum{n} \in \mathbb{R}^d$, and let $G = ([n],E)$ be a graph, where $[n] = \{1,2\ldots,n\}$, and $E = E_g \sqcup \Eb$, with $\Eb$ and $E_g$ corresponding to pairwise direction observations that are respectively corrupted and uncorrupted.  The uncorrupted observations are assumed to be noiseless.  That is, for each $ij \in E$, we are given a vector $\vij$, where

\begin{equation} \label{measurements}
\begin{array}{ll}
v_{ij} = \frac{\toi - \toj}{\bigl \|\toi - \toj \bigr\|_2}   \text{ for $(i,j)$ in $E_g$}\\
v_{ij} \in S^{d-1} \text{ arbitrary, for $(i,j) \in \Eb$}.\\
\end{array}
\end{equation}




Consider the task of recovering the locations $T$ up to a global translation and scale, from only the observations $\{ v_{ij} \}_{ij \in E}$, and without any knowledge about the decomposition $E = E_g  \sqcup E_b $, nor the nature of the pairwise direction corruptions. For $d=3$, this problem corresponds to (iii) once an estimate of directions is provided. 
The location recovery problem is to recover a set of points in $\mathbb{R}^d$ from observations of pairwise directions between those points.  
Since relative direction observations are invariant under a global translation and scaling, one can at best hope to recover the locations $\Tnot = \{ \tonum{1},\ldots, \tonum{n}\}$ up to such a gauge transformation. That is, successful recovery from $\{v_{ij}\}_{(i,j) \in E}$ is finding a set of vectors ${\{\alpha(\toi + w)\}_{i \in [n]}}$ for some $w \in \mathbb{R}^d$ and $\alpha >0$. We will say that two sets of n vectors $T = \{t_1,\ldots,t_n\}$ and $\Tnot$ are equal up to global translation and scale if there exists a vector $w$ and a scalar $\alpha >0 $ such that $t_i = \alpha(\toi + w)$ for all $i \in [n]$.  In this case, we will say that $T \sim \Tnot$. The location recovery problem is then stated as:
\begin{alignat}{3} \label{formulation}
&\text{Given:} &\quad &G([n],E), \quad \{\vij\}_{ij \in E} \text{\ \  satisfying \eqref{measurements} }\notag\\
&\text{Find:} && T = \{t_1, \ldots, t_n \} \in \mathbb{R}^{d \times n}, \quad \text{such that} \quad T \sim \Tnot. 
\end{alignat}




Formally, let $\deg_b(i)$ be the degree of location $i$ in the graph $([n], \Eb)$ and note that we do not assume anything about the nature of corruptions. That is, we work with adversarially chosen corrupted edges $E_b$ and arbitrary corruptions of observations associated to those edges. To solve the location recovery problem in this challenging setting, we introduce a simple convex program called ShapeFit: 
\begin{equation} \label{ShapeFit}
  \begin{array}{ll}
    \smash{\displaystyle\min_{ t_i \in \mathbb{R}^3, i \in [n] }}   & \sum_{ij \in E} \| P_{v_{ij}^\perp} (t_i - t_j) \|_2\\
   \hskip 14 pt \text{s.t.} & \sum_{ij \in E} \langle t_i - t_j, v_{ij} \rangle = 1, \quad  \sum_{i=1}^n t_i = 0 \quad  \\
\end{array}
\end{equation}
where $\Pvijp$ is the projector onto the orthogonal complement of the span of $\vij$.  The objective in \eqref{ShapeFit} is robust to outliers because it has the structure of an $\ell_1$ norm of a set of unsquared distances. The constraints act to remove the scale and translational ambiguities.

This convex program is a second order cone problem with $dn$ variables and two constraints.  Hence, the search space has dimension $dn-2$, which is minimal due to the $dn$ degrees of freedom in the locations $\{t_i\}$ and the two inherent degeneracies of translation and scale.

\subsection{Theoretical guarantees and practical implications}

Although we have established a much broader class of results w.r.t the assumptions on locations (see Appendix), we consider here the physically relevant and simple model where pairwise direction observations about $n$  i.i.d. Gaussian camera locations in $\mathbb{R}^3$ are given according to an \erdosrenyi random graph $G(n,p)$, which is a graph on $n$ vertices with each pair of vertices $(i,j)$ having an edge with probability $p$, independently of all other edges. In this setting, ShapeFit \eqref{ShapeFit} achieves exact recovery for any sufficiently large number of locations, provided that a poly-logarithmically small fraction of observations are adversarially corrupted at each node. 

\sloppy
\begin{theorem} \label{thm-3d} 
Let $G([n],E)$ be a random graph in $G(n,p)$ with\footnote{$p = \Omega(f(n))$ means that there exists a universal constant $C$ such that $p \geq C f(n)$} $p = \Omega( n^{-1/5} \log^{3/5} n.)$. Choose the locations of the vertices $ \tnot_1, \ldots \tnot_n  \in \R^3$ to be  i.i.d., independent vectors from the random normal distribution $\mathcal{N}(0, I_{3 \times 3}),$ and measure the pairwise directions $\vij \in \mathbb{S}^2$ between adjacent vertices.
Choose an arbitrary subgraph $\Eb$ satisfying $\max_i \deg_b(i) \leq \gamma n$ for some positive $\gamma.$  Corrupt these pairwise directions by applying an arbitrary modification to $\vij \in \mathbb{S}^2$ for $ij \in \Eb$.\\
\indent For $\gamma = \Omega(p^5 / \log^3 n)$ and sufficiently large $n,$ ShapeFit achieves exact recovery with high probability.   More precisely, with probability at least  $1- \frac{1}{n^{4}},$  the convex program \eqref{ShapeFit} has a unique minimizer equal to $\left \{\alpha \Bigl(\toi - \tnotbar \Bigr)\right\}_{i \in [n]}$ for some positive $\alpha$ and for $\tnotbar = \frac{1}{n}\sum_{i\in [n]} \toi$. 
\end{theorem}
\fussy

%

That is, provided the locations are i.i.d Gaussian and the underlying graph of observations is \erdosrenyi\!\!, ShapeFit is exact with high probability simultaneously for all corruption subgraphs of bounded degree with adversarially corrupted directions. To the best of our knowledge, our algorithms are the first to rest on theoretical results guaranteeing location recovery in the challenging case of corrupted pairwise direction observations.





The above result gives us confidence of the robustness of the method we propose, which is validated empirically in Sect. \ref{sect-expm} and supported theoretically in the Supplementary Material. Our main contribution in this paper is the design of efficient implementations based on the theory. Indeed, our empirical assessment shows that we can improve computational efficiency by one order of magnitude at accuracy roughly equal to existing state of the art methods. 

The efficiency and robustness of our method suggests its use as an alternative to the standard SfM pipeline for real-time applications, by replacing camera-to-camera direction estimation and triangulation with a single corruption-robust simultaneous recovery of camera locations and 3D structure - that is, by compressing two steps of the usual pipeline (ii)-(iii) into a single robust inference step.  This alternative applies to the case where rotations are known, for example through inertial measurements. This transforms the location recovery problem, where both camera locations and 3D points are represented as nodes in the graph, into a variant where the graph is bipartite, with edges only between camera positions and the 3D structure points. 
In the Supplementary Material, we present experimental results for the bipartite case.


\subsection{ Proof outline of Theorem 1}

A complete proof of Theorem 1 is involved and, and is included in the Appendix. A rough proof outline is as follows. Consider the true locations $t_1^{(0)}, \ldots t_n^{(0)}$ and a feasible perturbation $t_i^{(0)} + h_i$. For any $(i,j) \in E_g$, the objective increases from zero to $\|P_{(t_i^{(0)}-t_j^{(0)})^\perp}(h_i - h_j)\|_2$, while for $(k,l) \in E_b$ the objective may decrease as much as $\|h_k-h_l \|_2$. Optimality thus requires
\[
\sum_{ (i,j) \in E_g } \|P_{(t_i^{(0)}-t_j^{(0)})^\perp}(h_i - h_j)\|_2 > \sum_{(k,l) \in E_b} \|h_k- h_l\|_2.  
\]
We call, for any $(i,j) \in E$, $ \| P_{ t_{ij}^{(0)}}( h_i - h_j )\|_2$ and $ \| P_{ (t_{ij}^{(0)})^\perp}( h_i - h_j )\|_2$ the parallel and orthogonal deviation of $h_i-h_j$, respectively, and show
separately that (i) orthogonal and (ii) parallel deviation of bad edges induces sufficient orthogonal deviation on the good edges.  




The proof strategy in both cases is combinatorial propagation of a local geometric property. For case (i) we establish that if a collection of triangles in $\mathbb{R}^3$ share the same base and the locations opposite the base are sufficiently ``well-distributed,'' then an infinitesimal rotation of the base induces infinitesimal rotations in edges of many of the triangles. Then, for each corrupted edge $(k,l)$ we ensure that we can find sufficiently many triangles in the observation graph with two good edges and base $(k,l)$, with locations at the opposing vertices being ``well-distributed.'' Case (ii) is more nuanced and requires strongly using the constraints of the ShapeFit program. Here the local property is that for a tetrahedron in $\mathbb{R}^3$ with well distributed vertices, any discordant parallel deviations on two disjoint edges induce enough infinitesimal rotational motion on some other edge of the tetrahedron. Combinatorial propagation is then handled in two regimes of the relative balance of parallel deviations on the good and corrupted subgraphs.

\section{Numerical Approach} \label{sect:numerics}
We now study the efficient numerical minimization of the problem \eqref{ShapeFit} via the alternating direction method of multipliers (ADMM).  The ADMM approach is advantageous because each sub-step of the algorithm is efficiently solvable in closed form.  Also, unlike simple gradient methods, the ADMM method does not require smoothing/regularization of the $\ell_2$ norm penalty that results in poor conditioning. 
Problem \eqref{ShapeFit} can be reformulated as 
\begin{equation}   
  \begin{array}{ll} \label{constrained}
    \smash{\displaystyle\min_{ t \in \mathcal G}}   &  \quad\sum_{ij \in E} \| P_{v_{ij}^\perp} (y_{ij}) \|_2\\
   \hskip 6 pt \text{s.t.} &  \quad y_{ij} = t_i - t_j , \forall  \ij \in E.
\end{array}
\end{equation}
For notational simplicity, we have removed the constraints on $\{t_i\}$ and written the problem as a minimization over the set of all gauge-normalized point clouds in  $\mathbb{R}^3,$ denoted
 $$\mathcal G=\{T \in \mathbb{R}^{n\times 3} |\sum_{ij \in E} \langle t_i - t_j, v_{ij} \rangle = 1, \quad  \sum_{i=1}^n t_i = 0 \}$$
where $T=\{t_i\}$ is a collection containing $n$ vectors in $\mathbb R^3$.

  To derive an ADMM method, we now write the (scaled) augmented Lagrangian for \eqref{constrained}, which is \cite{boyd2011distributed,goldstein2014fast}
\begin{equation}   
  \begin{array}{ll} \label{constrained}
      \mathcal L_\rho(T,Y, \lambda)=& \sum_{ij \in E} \| P_{v_{ij}^\perp} (y_{ij}) \|_2 
        +\frac{\tau}{\rho}  \sum_{ij \in E} \|   t_i - t_j - y_{ij}  + \lambda_{ij}\|^2, 
\end{array}
\end{equation}
where $\rho$ is a constant stepsize parameter, and $\lambda = \{ \lambda_{ij}\}$ contains Lagrange multipliers. The solution to the constrained problem \eqref{constrained} corresponds to a saddle point of the augmented Lagrangian that is minimal for $T$ and $Y$ while being maximal with respect to $\lambda.$   ADMM finds this saddle point by iteratively minimizing $ \mathcal L_\rho(T,Y, \lambda)$ for $T$ and $Y,$ and then using a gradient ascent step to maximize for $\lambda.$  The corresponding updates are
  $$
  \begin{cases}
    T    \gets {\displaystyle \argmin_{ T \in \mathcal G}}    \,\, \mathcal L_\rho(T,Y, \lambda) \\
     Y \gets  {\displaystyle \argmin_{ Y \in \mathbb R^{|E|\times 3}}}  \,\,  \mathcal L_\rho(T,Y, \lambda)\\
     \lambda_{ij} \gets \lambda_{ij}+ t_i - t_j - y_{ij} .
       \end{cases}
$$
The minimization for $T$ is simply a least-squares problem. Let $R: \mathbb R^{n\times 3} \to \mathbb R^{|E|\times 3}$ be a linear operator such that the $k$th row of $RT$ is $t_i-t_j,$ where $(i,j)$ is the $k$th edge in $E$.  The $T$ update now has the form 
$$T \gets \argmin_{T\in \mathcal G} \| RT -Y+\lambda \|^2.$$ 
The solution to this minimization is found simply by computing the (possibly sparse) factorization of $R,$ and applying a rank-1 update (i.e., using the Sherman-Morrison formula) to account for the linear constraints in $\mathcal G.$ 

  We now examine the update for $y.$  Let $z_{ij}=t_i - t_j   + \lambda_{ij}.$  The updated value of $y_{ij}$  is then the minimizer of 
  \begin{align*}
   \|P_{v_{ij}^\perp}&(y_{ij})\| + \frac{\rho}{2}\|z_{ij}-y_{ij}\|^2 =  \|P_{v_{ij}^\perp}(y_{ij})\| \\
       &+ \frac{\rho}{2}\|P_{v_{ij}^\perp}(z_{ij}-y_{ij})\|^2 +\frac{\rho}{2}\|P_{v_{ij}}(z_{ij}-y_{ij})\|^2.
  \end{align*}
 The minimum of this objective has the closed form\footnote{shrink$(x, \lambda) = \text{sign}(x) \max(0,|x| - \lambda)$}.
  \begin{equation}
\label{ShapeKick}
y_{ij} \gets   P_{v_{ij}} (z_{ij})+ \shrink(P_{v_{ij}^\perp} (z_{ij}), 1/\rho).
\end{equation}  
  
Finally, it is known that the convergence of ADMM for 1-homogenous problems is (empirically) very fast for the first few iterations, and then convergence slows down.  For real-time applications, one may prefer a more aggressive algorithm that does not suffer from this slowdown.   Slowdown is often combated using ÒkickingÓ [24], and we adopt a variant of this trick to accelerate ShapeFit.  The kicking procedure starts with a small value of $\tau,$ and iterates until convergence stagnates (the values of $y$ become nearly constant).  We then increase $\tau$ by a factor of 10, and run the algorithm until it slows down again.  The ShapeKick approach drastically reduces runtime when moderate accuracy is needed, however it generally produces higher numerical errors than simply applying the un-kicked ADMM for a very long period of time.  

\section{Numerical Experiments}
\label{sect-expm}

For empirical validation we adopt here the data and protocol of the most common benchmarks for SfM and location recovery. We first verify that when the data is generated according to a model that satisfies the assumptions of the analysis, we indeed witness exact recovery despite a large fraction of corruptions. We then report representative results on benchmark datasets in a variety of experimental settings, where in some cases the assumptions may be violated. Although there is considerable performance variability among different methods on different datasets and no uniform winner, our scheme is competitive with the state-of-the-art in terms of accuracy, but at a fraction of the computational cost. The results are summarized in Sect. \ref{sect-summary}.

\subsection{Experiments on Synthetic Data}


\begin{figure} \label{phase}
\begin{center}
%
%
\begin{psfrags}%
\psfragscanon%
%
\psfrag{s03}[b][b]{\color[rgb]{0,0,0}\setlength{\tabcolsep}{0pt}\begin{tabular}{c}LUD, noise\end{tabular}}%
\psfrag{s04}[b][b]{\color[rgb]{0,0,0}\setlength{\tabcolsep}{0pt}\begin{tabular}{c}ShapeFit, noise\end{tabular}}%
\psfrag{s05}[t][t]{\color[rgb]{0.15,0.15,0.15}\setlength{\tabcolsep}{0pt}\begin{tabular}{c}$p$\end{tabular}}%
\psfrag{s06}[b][b]{\color[rgb]{0,0,0}\setlength{\tabcolsep}{0pt}\begin{tabular}{c}ShapeFit, no noise\end{tabular}}%
\psfrag{s07}[b][b]{\color[rgb]{0,0,0}\setlength{\tabcolsep}{0pt}\begin{tabular}{c}LUD, no noise\end{tabular}}%
\psfrag{s08}[b][b]{\color[rgb]{0.15,0.15,0.15}\setlength{\tabcolsep}{0pt}\begin{tabular}{c}$q$\end{tabular}}%
\psfrag{s11}[t][t]{\color[rgb]{0.15,0.15,0.15}\setlength{\tabcolsep}{0pt}\begin{tabular}{c}$p$\end{tabular}}%
\psfrag{s12}[b][b]{\color[rgb]{0.15,0.15,0.15}\setlength{\tabcolsep}{0pt}\begin{tabular}{c}$q$\end{tabular}}%
%
\color[rgb]{0.15,0.15,0.15}%
%
\psfrag{x01}[t][t]{0.3}%
\psfrag{x02}[t][t]{0.6}%
\psfrag{x03}[t][t]{0.9}%
\psfrag{x04}[t][t]{0.3}%
\psfrag{x05}[t][t]{0.6}%
\psfrag{x06}[t][t]{0.9}%
\psfrag{x07}[t][t]{0.3}%
\psfrag{x08}[t][t]{0.6}%
\psfrag{x09}[t][t]{0.9}%
\psfrag{x10}[t][t]{0.3}%
\psfrag{x11}[t][t]{0.6}%
\psfrag{x12}[t][t]{0.9}%
%
\psfrag{v01}[r][r]{0.75}%
\psfrag{v02}[r][r]{0.5}%
\psfrag{v03}[r][r]{0.25}%
\psfrag{v04}[r][r]{0}%
\psfrag{v05}[r][r]{0.75}%
\psfrag{v06}[r][r]{0.5}%
\psfrag{v07}[r][r]{0.25}%
\psfrag{v08}[r][r]{0}%
\psfrag{v09}[r][r]{0.75}%
\psfrag{v10}[r][r]{0.5}%
\psfrag{v11}[r][r]{0.25}%
\psfrag{v12}[r][r]{0}%
\psfrag{v13}[r][r]{0.75}%
\psfrag{v14}[r][r]{0.5}%
\psfrag{v15}[r][r]{0.25}%
\psfrag{v16}[r][r]{0}%

\psfrag{b1}[l][l]{0}%
\psfrag{b2}[l][l]{0.25}%
\psfrag{b3}[l][l]{0.5}%
\psfrag{a1}[l][l]{$10^{-10}$}%
\psfrag{a2}[l][l]{$10^{-5}$}%
\psfrag{a3}[l][l]{$10^{0}$}%
\resizebox{11cm}{!}{\includegraphics{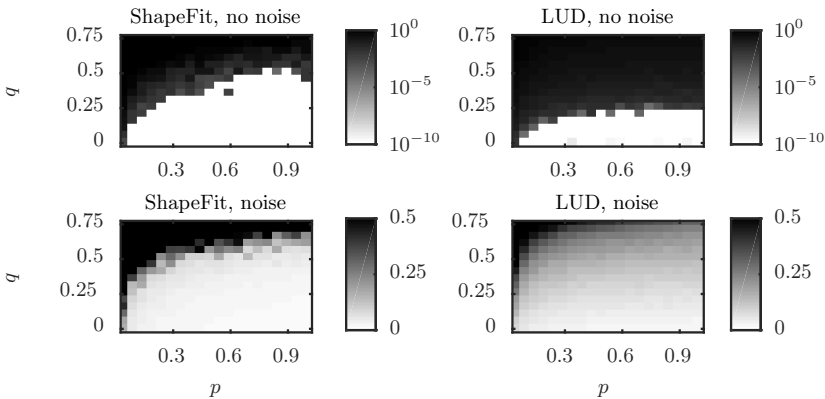}}%
\end{psfrags}%
%

\end{center}
   \caption{RFE (\ref{RFE}) results for ShapeFit and LUD on synthetic noiseless + corrupted and noisy + corrupted data. The grayscale intensity of each pixel corresponds to average RFE over 10 random trials, depending on the edge probability $p$ and corruption probability $q$. 
 Direction observations are generated by equation (\ref{noise model}) with $\sigma = 0$ for the top two tables and with $\sigma = 0.05$ for the bottom two tables.  
}
    \label{phase}
\end{figure}



In this section we validate ShapeFit on synthetic data and compare its performance with that of the LUD algorithm of \cite{AMIT}, with both algorithms implemented in our ADMM framework. In particular we report on ShapeFit's \emph{exact} location recovery from partially corrupted pair-wise directions and stable recovery from noisy and partially corrupted directions. The LUD method also exhibits both of these phenomena, and we compare the (empirical) phase transition diagrams of both methods in identical regimes. 

The locations $\{t_i \}_{i=1}^n$ to be recovered are i.i.d $\mathcal{N}(0,I_{3\times 3})$. The graph of pair-wise observations $G([n],E)$ is drawn independently from the \erdosrenyi model $\mathcal{G}(n,p)$, that is each edge $(i,j)$ is in $E$ with probability $p$, independently from all other edges. Having drawn locations $\{t_i \}_{i=1}^n$ and $G([n],E)$, consider i.i.d random variables $\eta_{ij} =^d \mathcal{N}(0,I_{3 \times 3})$ for $(i,j) \in E$ independent from all other random variables and let 
\vspace{-3 pt}
\begin{equation} \label{noise model}
 \tilde{v}_{ij} =
  \begin{cases} 
      \hfill  \eta_{ij}   \hfill & \text{with probability $q$}  \\
      \hfill \frac{t_i -t_j}{\|t_i - t_j\|_2} + \sigma \eta_{ij} \hfill & \text{ otherwise,} \\
  \end{cases}
\end{equation}
where $\sigma \geq 0$ controls the noise level and the assignments are made independently on each edge in $E$. We then obtain pair-wise direction observations as $v_{ij} = \frac{\tilde{v}_{ij}}{\|\tilde{v}_{ij}\|_2}$ for each $(i,j) \in E$, and thus $v_{ij}$ is a random direction on the unit sphere with corruption probability $q$ and is a noisy version of the true pair-wise direction with probability $1-q$. 


\begin{figure} \label{phase_noise}
\begin{center}
%
%
\begin{psfrags}%
\psfragscanon%
%
\psfrag{s01}[t][t]{\color[rgb]{0.15,0.15,0.15}\setlength{\tabcolsep}{0pt}\begin{tabular}{c}$\sigma$\end{tabular}}%
\psfrag{s02}[b][b]{\color[rgb]{0.15,0.15,0.15}\setlength{\tabcolsep}{0pt}\begin{tabular}{c}Average RFE\end{tabular}}%
\psfrag{s03}[b][b]{\color[rgb]{0,0,0}\setlength{\tabcolsep}{0pt}\begin{tabular}{c}$n=200$, $p=0.50$, $q=0.3$\end{tabular}}%
\psfrag{s04}[b][b]{\color[rgb]{0.15,0.15,0.15}\setlength{\tabcolsep}{0pt}\begin{tabular}{c}Average RFE\end{tabular}}%
\psfrag{s05}[t][t]{\color[rgb]{0.15,0.15,0.15}\setlength{\tabcolsep}{0pt}\begin{tabular}{c}$\sigma$\end{tabular}}%
\psfrag{s06}[b][b]{\color[rgb]{0,0,0}\setlength{\tabcolsep}{0pt}\begin{tabular}{c}$n=200$, $p=0.25$, $q=0.1$\end{tabular}}%
%
\color[rgb]{0.15,0.15,0.15}%
%
\psfrag{x01}[t][t]{$10^{-10}$}%
\psfrag{x02}[t][t]{$10^{-5\ }\hspace{0.2em}$}%
\psfrag{x03}[t][t]{$10^{0\ \ \ }\hspace{0.2em}$}%
\psfrag{x04}[t][t]{$10^{-10}$}%
\psfrag{x05}[t][t]{$10^{-5\ }\hspace{0.2em}$}%
\psfrag{x06}[t][t]{$10^{0\ \ \ }\hspace{0.2em}$}%
%
\psfrag{v01}[r][r]{$10^{-10}$}%
\psfrag{v02}[r][r]{$10^{-5\ }\hspace{0.2em}$}%
\psfrag{v03}[r][r]{$10^{0\ \ \ }\hspace{0.2em}$}%
\psfrag{v04}[r][r]{$10^{-10}$}%
\psfrag{v05}[r][r]{$10^{-5\ }\hspace{0.2em}$}%
\psfrag{v06}[r][r]{$10^{0\ \ \ }\hspace{0.2em}$}%
%
\resizebox{12cm}{!}{\includegraphics{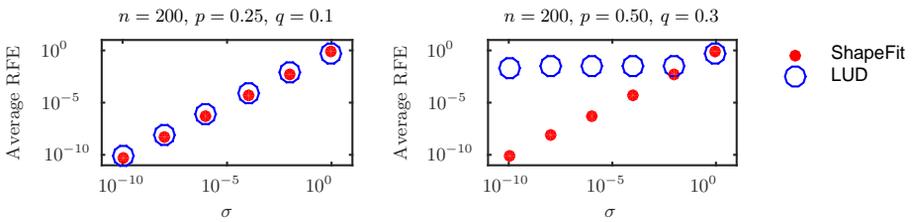}}%
\end{psfrags}%
%

\end{center}
  \caption{Mean RFE for ShapeFit and LUD on synthetic data, as a function of the noise parameter $\sigma$.}
\label{phase_noise}

\end{figure}


We evaluate recovery performance in terms of a relative Frobenius error (RFE). For any set of locations $\{x_i\}_{i=1}^n$ in $\mathbb{R}^d$, let $T(x_1,\ldots,x_n)$ be a $d \times n$ matrix with $i^{\text{th}}$ column given by $x_i - \sum_{i=1}^n x_i$. Define $T_0 = T(t_1,\ldots,t_n)$ as the matrix of original locations and let $\{\hat{t}_i\}_{i=1}^n$ be the set of recovered locations.  Define $\hat{T} = T(\hat{t}_1,\ldots,\hat{t}_n)$. Then the RFE is given by 
\begin{equation} \label{RFE}
\text{RFE}(T_0,\hat{T}) = \left\| T_0/\|T_0\|_F -\hat{T}/\|\hat{T}\|_F \right\|_F,
\end{equation}
 which accounts for the global translation and scale ambiguity, where $\| .\|_F$ is the Frobenius norm on $\mathbb{R}^{d \times n}$.  Note that an RFE of zero corresponds to exact recovery.

For each pixel in the phase diagrams of Figure (\ref{phase}), we generate 10 independent random recovery problems as described above, recover locations using ShapeFit (left column) or LUD (right column), and record the average RFE as a grayscale intensity. The first set of experiments considers recovery from partially corrupted and otherwise noiseless ($\sigma =0$) directions. We note that in the top row of phase diagrams for both methods, we see $\emph{exact}$ recovery from partially corrupted direction observations (we define exact recovery as RFE $ < 10^{-9})$. ShapeFit has a wider region of exact recovery in the $(p,q)$ parameter space, exhibiting exact recovery at up to between $10 \%$ and $50 \%$ of corruption (depending on $p$ and $n$), while LUD stops being exact at around $20 \%$ corruption. 

The second set of experiments considers recovery from partially corrupted and otherwise noisy ($\sigma > 0)$ directions. We take $\sigma = 0.05$ to generate the bottom row of tables and consider phase transitions on a coarser scale of RFE. We see that recovery is stable from noisy and partially corrupted direction observations for both ShapeFit and LUD, with ShapeFit having a more favorable recovery profile at the lower range of corruptions in that the recovery is more accurate than LUD up to the rapid phase transition, while LUD's performance starts to degrade at a lower level of corruption yet continues to provide meaningful recovery slightly above the level of corruption of ShapeFit's phase transition. 

In Figure (\ref{phase_noise}) we provide further numerical experiments that illustrate that ShapeFit and LUD have graceful degradation of recovery with respect to noise.

\subsection{Setup of Experiments on Real Data}
\label{sect-setup}

We validate our method ShapeFit (\ref{ShapeFit}) on 13 benchmark datasets containing irregular collections of images of real scenes from \cite{1dsfm}. We  compare its performance to that of LUD \cite{AMIT} and 1dSfM. We implement two fast versions of ShapeFit and a fast version of LUD based on the Alternating Directions Multiplier Method (ADMM) and an aggressive step-size selection method. We refer to the faster of the two implementations of ShapeFit as ShapeKick. To solve 1dsfm we use code provided by \cite{1dsfm}. We perform our experiments on an Intel(R) Core (TM) i5 CPU with 2 cores, running at 2.6 GHz. A unique aspect of our numerical comparisons is that we run ShapeFit, ShapeKick, LUD, and 1dSfM on the same problem instances generated from several different regimes.  Thus, we measure head-to-head performance on the location recovery task objectively. We emphasize that we do no dataset-specific tuning of the recovery algorithms. For each problem instance, we report on the median and mean Euclidean distance error between estimated camera locations and the ground truth for each algorithm, as in \cite{1dsfm,AMIT}.

To generate problem instances for each dataset, we first solve for global camera rotations using the method of Govindu \cite{1dSfM_5}, then solve for relative directions between cameras using epipolar geometry, and obtain rotation estimates using code provided by Snavely and Wilson \cite{1dsfm} for both of these steps. After this step, we have obtained directions among cameras, directions between cameras, and 3d structure points, all in the same reference frame. Let $G_s([n] \times [m], E_c)$ be the obtained bipartite graph of directions between the $n$ camera locations and $m$ structure points (where we associate the appropriate direction to each edge), and similarly let $G_c([n],E_l)$ be the obtained graph of directions between the $n$ camera locations. To generate problem instances, we consider directions computed as functions of $G_s \sqcup G_c$. 

The first problem instance regime is that of using robust PCA to re-compute pairwise direction estimates between cameras, as used by Singer and Ozyesil in \cite{AMIT}. We use code provided in \cite{AMIT} and refer to this regime as Robust PCA. We also generate problem instances using the greedy pruning technique used by \cite{1dsfm}, which proceeds by selecting a subset $G_{s}^{(k)}$ of $G_s$ greedily to ensure that each pair of selected cameras have at least $k$ co-visible structure points via edges in $G_{s}^{(k)}$ where $k$ is an integer parameter. We use code provided by Snavely and Wilson to generate these subgraphs $G_{s}^{(k)}$ of camera-to-structure directions for $k=6$ and $k=50$ \cite{1dsfm}. We consider $G_{s}^{(k)} \sqcup G_c$ as the resulting two problem instances, referred to as Monopartite $k=6$ and $k=50$. These Monopartite problem instances are exactly the same as those generated in \cite{1dsfm}. Finally, we consider purely bipartite versions of these problems, by keeping just the camera-to-structure directions $G_{s}^{(k)}$ for $k=6$ and $k=50$ and ignoring translation estimates from epipolar geometry.  We refer to these instances as Bipartite $k=6$ and $k=50$. Thus, the bipartite problem instances are strict subsets of the monopartite instances and do not require any epipolar geometry to set up aside from global rotation estimation. In sum, this gives five problem instances per dataset.

As in \cite{1dsfm,AMIT}, we consider the ground truth as camera location estimates provided by a sequential SfM solver provided by Snavely and Wilson. To compute the global translation and scale between recovered solutions and the ground truth we use a RANSAC-based method as in \cite{1dsfm}, using their code.

Table 1 shows the median and mean reconstruction errors (without bundle adjustment) for seven recovery algorithms on thirteen datasets under two monopartite problem formulations.  Table 2 reports runtimes needed by different methods to set up and solve translation problems. Table 3 in the supplementary material shows the reconstruction errors under three additional problem formulations, including both bipartite formulations. Table 4 in the supplementary material shows the runtimes for these additional problem formulations.  In Table 1, the best median error (among all algorithms) for each dataset and formulation is marked in bold.  The best median error (among all algorithms and all five formulations) is marked in red and with an asterisk.  The seven algorithms considered are: ShapeKick, ShapeFit, LUD, 1dSfM outlier removal followed by nonlinear least squares solver, 1dSfM followed by a Huber loss solver, 1dSfM followed by ShapeKick, and 1dSfM followed by LUD.  The recovery errors are relative to the estimates from \cite{1dsfm}, which were computed by a sequential SfM solver.  Ties are resolved by less significant digits not displayed.

\begin{table}[h!] \label{errors}
\begin{center}
\begin{tabular}{| l | c |c || c|c || c|c || c|c || c|c || c|c || c|c || c|c ||}
\cline{4-17}
\multicolumn{3}{c|}{}& \multicolumn{6}{c||}{without 1dSfM} & \multicolumn{8}{c||}{with 1dSfM}\\\hline
Dataset & \multicolumn{2}{c||}{Size} & \multicolumn{2}{|c ||}{SK} & \multicolumn{2}{|c ||}{SF} & \multicolumn{2}{|c ||}{LUD} & \multicolumn{2}{|c ||}{NLS} & \multicolumn{2}{|c ||}{Huber} & \multicolumn{2}{|c ||}{SK} & \multicolumn{2}{|c ||}{LUD}\\ \cline{2-17}
& $N_c$ & $N_\ell$ & $\tilde{e}$ & $\hat{e}$ & $\tilde{e}$ & $\hat{e}$ & $\tilde{e}$ & $\hat{e}$ & $\tilde{e}$ & $\hat{e}$ & $\tilde{e}$ & $\hat{e}$ & $\tilde{e}$ & $\hat{e}$ &  $\tilde{e}$ & $\hat{e}$ \\  \hline 
\multicolumn{10}{l}{\hspace{1em} Monopartite $k=6$ formulation:} \\\hline
Ellis Island & 227 & 365 & 2.7 & 380& 5.7 & 15& 4.1 & 9.8& 3.4 & 10& \bf{{\color{red} 1.7$*$}} & 8.9& 1.9 & 12& 3.5 & 9.7\\ 
NYC Library & 332 & 706 & 4.4 & 186& 3.7 & 194& 2.0 & 4.2& 1.8 & 738& \bf{{\color{red} 1.0$*$}} & 5e3& 1.4 & 162& 1.9 & 5.0\\ 
Piazza Pop. & 338 & 558 & \bf{2.4} & 8.5& 3.6 & 138& 4.0 & 6.1& 3.1 & 156& 3.3 & 19& 3.6 & 5.9& 3.9 & 6.2\\ 
Metropolis & 341 & 686 & 25 & 979& 10 & 80& 6.3 & 16& 8.1 & 7e3& \bf{4.0} & 1e4& 6.0 & 81& 6.4 & 16\\ 
Montreal ND & 450 & 728 & 1.4 & 2.7& 1.4 & 3.5& 0.8 & 1.4& 1.3 & 514& 0.8 & 2e3& 0.8 & 1.7& \bf{0.6} & 1.2\\ 
Tow. London & 472 & 914 & 3.9 & 2e3& 17 & 752& 8.4 & 25& 11 & 401& 2.8 & 8e4& \bf{{\color{red} 2.3$*$}} & 164& 7.8 & 24\\ 
Notre Dame & 553 & 726 & 0.4 & 3.6& 0.3 & 4.0& 1.6 & 4.0& 0.9 & 669& 0.2 & 5e3& \bf{{\color{red} 0.2$*$}} & 1.5& 2.3 & 3.5\\ 
Alamo & 577 & 950 & 1.0 & 4.1& 2.5 & 6.0& 2.4 & 4.5& 1.0 & 6e3& \bf{0.8} & 2e3& 0.9 & 5.0& 1.6 & 3.5\\ 
Gendarmen. & 677 & 1165 & 52 & 111& \bf{32} & 487& 33 & 57& 50 & 2e3& 38 & 7e4& 53 & 236& 34 & 59\\ 
Union Sq. & 789 & 1660 & 10 & 123& 12 & 84& 5.5 & 12& 5.6 & 4e3& \bf{{\color{red} 4.9$*$}} & 5e3& 8.9 & 47& 5.0 & 11\\ 
Vienna Cath. & 836 & 1636 & 3.4 & 20& 11 & 5e3& 5.6 & 11& 5.1 & 4e3& 2.1 & 7e3& \bf{{\color{red} 1.9$*$}} & 11& 3.2 & 11\\ 
Roman For. & 1084 & 1786 & 39 & 2e3& 12 & 25& 12 & 23& 5.7 & 1e4& \bf{3.0} & 5e4& 4.3 & 25& 6.6 & 15\\ 
Piccadilly & 2152 & 3815 & 3.8 & 127& 3.7 & 122& 2.8 & 5.4& 2.5 & 800& 1.5 & 7e3& \bf{{\color{red} 1.2$*$}} & 15& 2.4 & 5.4\\  \hline
\multicolumn{10}{l}{\hspace{1em} Robust PCA  formulation:} \\\hline
Ellis Island & 227 & 245 & 30 & 442& 25 & 5e4& \bf{25} & 25& 32 & 3e3& 40 & 1e6& 29 & 1e4& 25 & 25\\ 
NYC Library & 332 & 370 & 2.5 & 3e3& 2.5 & 3e3& 2.9 & 7.2& 4.3 & 3e3& \bf{2.2} & 995& 2.4 & 9.9& 2.8 & 6.9\\ 
Piazza Pop. & 338 & 352 & 2.4 & 8.9& 1.8 & 96& 3.0 & 6.2& 2.6 & 3e3& 3.2 & 1e5& \bf{1.7} & 8.8& 2.0 & 6.5\\ 
Metropolis & 340 & 391 & 2.8 & 145& 7.9 & 2e5& 4.2 & 15& 7.8 & 3e4& 4.0 & 6e4& \bf{{\color{red} 2.4$*$}} & 73& 3.7 & 15\\ 
Montreal ND & 450 & 474 & 1.6 & 3.1& 1.7 & 3.8& 1.2 & 2.1& 1.1 & 2e4& \bf{0.9} & 4e4& 1.5 & 3.0& 1.1 & 1.9\\ 
Tow. London & 472 & 505 & 3.3 & 99& 3.4 & 510& 5.6 & 24& 16 & 6e4& 3.5 & 2e5& \bf{3.3} & 24& 4.3 & 22\\ 
Notre Dame & 553 & 553 & 0.5 & 1.5& 0.5 & 1.4& 0.5 & 1.5& 0.8 & 2e3& 0.5 & 5e3& 0.5 & 1.5& \bf{0.5} & 1.4\\ 
Alamo & 577 & 623 & 0.9 & 3.4& 0.9 & 41& 0.9 & 2.8& 0.9 & 7e3& 0.8 & 8e3& \bf{0.8} & 2.8& 0.9 & 2.6\\ 
Gendarmen. & 677 & 738 & 35 & 266& 33 & 5e3& 29 & 53& 36 & 1e4& 37 & 2e5& \bf{{\color{red} 27$*$}} & 152& 27 & 53\\ 
Union Sq. & 789 & 930 & 13 & 4e4& 9.1 & 1e4& 7.8 & 13& 9.4 & 5e3& 7.9 & 8e3& \bf{7.4} & 2e3& 7.9 & 13\\ 
Vienna Cath. & 836 & 915 & 19 & 2e3& 11 & 5e4& 6.0 & 15& 8.1 & 7e4& \bf{4.3} & 2e5& 7.6 & 70& 5.8 & 14\\ 
Roman For. & 1082 & 1126 & 18 & 661& 21 & 2e5& 7.6 & 18& 7.6 & 8e4& \bf{6.4} & 6e4& 19 & 166& 7.7 & 18\\ 
Piccadilly & 2151 & 2489 & 2.1 & 330& 4.4 & 8e3& 2.1 & 4.5& 2.9 & 5e3& \bf{1.8} & 3e4& 2.1 & 330& 2.1 & 4.6\\  \hline
\end{tabular}

\end{center}
\vspace{-2mm}
\caption{ Median ($\tilde{e}$) and mean ($\hat{e}$) reconstruction errors (in meters) across multiple datasets and problem formulations.  $N_c$ and $N_l$ denote the number of camera locations and the number of directions (camera-to-camera and camera-to-structure), respectively. The best performing algorithm in each row is bolded.  For each dataset, the best performing combination of algorithm and problem instance is starred with an asterisk.  }
 \label{errors}
\end{table}
\subsection{Summary and Analysis of Experiments on Real Data}
\label{sect-summary}

We observe that ShapeKick with 1dSfM outlier removal is a competitive method for location recovery from directions, as measured by median reconstruction error.  Table 1 shows that the combination of 1dSfM with ShapeKick has the smallest median reconstruction error\footnote{Six of these can be seen in Table 1, and two can be seen in the monopartite $k=50$ case in the Supplemental Materials.} for eight of the thirteen datasets.  The combination of 1dSfM and Huber has the smallest median error for three datasets.  ShapeKick without 1dSfM has the smallest median error for one dataset.  Finally, ShapeFit without 1dSfM has the smallest median error for one dataset (see Table 3 in the supplementary material).

We observe that ShapeKick is faster than previously published location recovery algorithms by a factor of 10-50.   Table 2 shows that in all cases, ShapeKick with or without 1dSfM are the fastest translations algorithms by wide margins.    ShapeKick with 1dSfM is typically slower than ShapeKick alone by up to a factor of two.  In a few cases, ShapeKick with 1dSfM is faster than ShapeKick alone because the outlier removal permits faster numerical convergence.   

We observe that ShapeKick can sometimes result in lower reconstruction errors than ShapeFit.  This effect is possible because the output of ShapeFit is not equal to ground truth.  Hence, the output of ShapeKick, which is an approximation of the output of ShapeFit, may return higher or lower reconstruction errors, especially after the outlier-tolerant RANSAC-based error estimation.

We observe that camera-to-camera measurements, though noisy and not directly measured, act to stabilize the location recovery problem in these phototourism datasets. Table 1 shows that the smallest median reconstruction error is achieved by the monopartite $k=6$ formulation for seven datasets.  The monopartite $k=50$ formulation is best for three datasets.  The Robust PCA formulation is best for two datasets.  Finally, the bipartite $k=6$ formulation is best for one dataset (see Table 3 in the supplementary material).  

\begin{table}[h!] \label{timings}
\begin{center}
\begin{tabular}{| l | c |c || c||c || c|| c|| c||c || c|||| c || c |}
\cline{4-10}
\multicolumn{3}{c|}{}& \multicolumn{3}{c||}{without 1dSfM} & \multicolumn{4}{c||||}{with 1dSfM}\\\hline
Dataset & $T_{\text{rot}}$ & $T_{\text{trans}}$ & SK & SF & LUD & NLS & Huber & SK & LUD & \cite{AMIT} & \cite{1dsfm}\\ \hline
\multicolumn{10}{l}{\hspace{1em} Monopartite $k=6$ formulation:} \\\hline
Ellis Island & 5.9 & 2.9 & \bf{0.6}& 7.2& 7.4& 33& 37& 1.4& 6.7& & 13\\ 
NYC Library & 7.3 & 8.6 & \bf{1.7}& 14& 14& 67& 26& 2.2& 10& & 54\\ 
Piazza Pop. & 11 & 4.6 & \bf{1.5}& 9.2& 11& 24& 115& 1.9& 8.6& & 35\\ 
Metropolis & 9.7 & 6.9 & \bf{1.2}& 9.0& 9.3& 58& 83& 2.4& 9.5& & 20\\ 
Montreal ND & 14 & 15 & \bf{2.4}& 28& 28& 60& 50& 3.5& 22& & 75\\ 
Tow. London & 6.6 & 15 & \bf{2.1}& 10& 11& 48& 43& 2.8& 10& & 55\\ 
Notre Dame & 38 & 23 & 7.5& 48& 14& 133& 66& \bf{7.1}& 17& & 59\\ 
Alamo & 41 & 16 & \bf{8.4}& 43& 41& 69& 202& 11& 37& & 73\\ 
Gendarmen. & 17 & 13 & \bf{3.5}& 20& 21& 60& 43& 4.8& 16& & \\ 
Union Sq. & 10 & 24 & \bf{2.0}& 17& 17& 48& 116& 3.7& 17& & 75\\ 
Vienna Cath. & 82 & 66 & \bf{4.4}& 52& 54& 436& 462& 8.2& 48& & 144\\ 
Roman For. & 28 & 52 & \bf{6.8}& 42& 44& 166& 130& 9.5& 28& & 135\\ 
Piccadilly & 826 & 424 & \bf{26}& 240& 204& 405& 593& 40& 163& & 366\\  \hline
\multicolumn{10}{l}{\hspace{1em} Robust PCA  formulation:} \\\hline
Ellis Island & 5.9 & 360 & \bf{0.5}& 5.9& 6.1& 3.2& 8.8& 1.3& 4.8& & \\ 
NYC Library & 7.3 & 906 & \bf{1.2}& 6.4& 6.5& 33& 38& 1.2& 5.3& 57& \\ 
Piazza Pop. & 11 & 314 & \bf{0.4}& 7.5& 2.8& 19& 7.6& 1.3& 6.8& 35& \\ 
Metropolis & 9.7 & 527 & \bf{0.9}& 6.6& 7.0& 36& 18& 1.7& 7.6& 27& \\ 
Montreal ND & 14 & 5e3 & \bf{1.3}& 24& 13& 1e4& 115& 3.4& 19& 112& \\ 
Tow. London & 6.6 & 2e3 & \bf{1.2}& 6.4& 6.8& 32& 142& 1.5& 7.4& 41& \\ 
Notre Dame & 38 & 2e4 & \bf{2.9}& 40& 24& 159& 46& 7.1& 32& 247& \\ 
Alamo & 41 & 3e3 & \bf{2.8}& 34& 18& 75& 199& 6.6& 41& 186& \\ 
Gendarmen. & 17 & 610 & \bf{1.8}& 17& 16& 70& 24& 3.3& 14& & \\ 
Union Sq. & 10 & 679 & \bf{1.6}& 10& 11& 52& 44& 2.6& 9.2& & \\ 
Vienna Cath. & 82 & 1e4 & \bf{6.8}& 41& 29& 283& 201& 6.8& 26& 255& \\ 
Roman For. & 28 & 5e3 & \bf{4.0}& 25& 24& 87& 82& 5.7& 21& & \\ 
Piccadilly & 826 & 4e3 & \bf{40}& 135& 143& 369& 364& \bf{40}& 182& & \\  \hline
\end{tabular}

\end{center}
\caption{Running times for the algorithms in seconds.  $T_\text{rot}$ and $T_\text{trans}$ provide the time to solve the rotations problem and to set up the translation problem, respectively.  Columns 4--10 present the times for solving the translations problem by our implementations of the respective algorithms.  
}
 \label{timings}
\end{table}

We observe that outlier filtering by 1dSfM enhances the outlier tolerance of convex methods like ShapeKick, ShapeFit, and LUD.  As an example, consider the Roman Forum dataset under a monopartite $k=6$ formulation.  The recovery errors of SK, SF, and LUD decrease by a factor or 2--10 by 1dSfM filtering.

We observe that outlier-robust location recovery methods are helpful even if 1dSfM is used to initially filter outliers.  Notice that for all reported simulations, the outlier-intolerant NLS algorithm never has the smallest median error.

Finally, we comment on the choice of the selection of mean and median recovery error as a metric.  Mean errors are susceptible to recovered locations that are outliers.  Median errors more accurately measure the overall shape of the set of locations.  Table 1 reveals that the LUD method has significantly lower mean reconstruction errors than any other method.  The mean reconstruction errors of ShapeKick, while higher than those of LUD, are still much smaller than those of 1dSfM with a Huber loss minimization.   Thus, LUD produces typically does not contain significant outliers, and ShapeFit contains outliers that are less significant than those from 1dSfM with a Huber loss.   





\section{Conclusion}
\label{sect-conclusion}
We propose a simple convex program called ShapeFit for location recovery, which comes with theoretical guarantees of exact location recovery from partially corrupted pairwise observations.  We propose a highly efficient numerical framework and use it to implement ShapeFit and LUD, producing runtime speedups of 10X or more over other implementations. Our fastest version of ShapeFit, called ShapeKick, is consistently at least 10X faster than previously published location recovery methods. We provide experiments on synthetic data illustrating exact recovery and stability of ShapeFit and LUD, and a thorough empirical comparison between ShapeFit, LUD, and 1dSfM on real data shows comparable reconstruction performance between the methods. 


We stress that our algorithm is the first to rely on provable performance guarantees despite adversarial corruptions. Such corruptions include photometric ambiguities due to repeated structures in man-made environments, a common occurrence in SfM. 
We have validated the results on synthetic datasets, as well as on public benchmarks, and demonstrated that ShapeFit achieves comparable reconstruction error to state of the art methods with a 10X speedup. 






\textbf{Acknowledgments.} TG was partially supported by National Science Foundation CCF-1535902 and by US Office of Naval Research grant N00014-15-1-2676.  PH was partially supported by National Science Foundation DMS-1464525.  CL was partially supported by the National Science Foundation DMS-1362326.  SS was partially supported by Air Force Office of Scientific Research FA9550-15-1-0229. VV was partially supported by the Office of Naval Research.

\clearpage

\bibliographystyle{splncs03}
\bibliography{references}
\end{document}